%% file: arxiv_main.tex
\documentclass[10pt,twocolumn,letterpaper]{article}

\usepackage[pagenumbers]{cvpr}      

\input{preamble}

\definecolor{cvprblue}{rgb}{0.21,0.49,0.74}

\def\paperID{1426} 
\def\confName{CVPR}
\def\confYear{2025}

\title{\methodname: What Could Happen Next?}

\author{Karran Pandey$^1$
\and
Matheus Gadelha$^2$
\and
Yannick Hold-Geoffroy$^2$
\and
Karan Singh$^1$
\and
Niloy J. Mitra$^{2,3}$
\and
Paul Guerrero$^2$
}
\begin{document}
\twocolumn[{%
\renewcommand\twocolumn[1][]{#1}%
\maketitle
\vspace{-25pt}
\begin{minipage}{\textwidth}
\centering
{\large $^1$ University of Toronto \qquad $^2$ Adobe Research \qquad $^3$ UCL}
\end{minipage}

\vspace{10pt}

\begin{minipage}{\textwidth}
\centering
\url{https://motionmodes.github.io}
\end{minipage}

\vspace{20pt}

}]
\input{sec/0_abstract}    
\input{sec/1_intro}

\input{sec/2_related_work}

\input{sec/3_method}
\input{sec/4_results}
\input{sec/5_discussion}

{
    \small
    \bibliographystyle{ieeenat_fullname}
    \bibliography{main}
}

\appendix

\clearpage

\input{sec/X_suppl}

\end{document}

%% file: preamble.tex
\usepackage{microtype}
\usepackage{mathtools}
\usepackage{xspace}
\usepackage{epigraph}
\usepackage{tabularx}
\usepackage{microtype}

\newcommand{\methodname}{Motion Modes\xspace}
\newcommand{\motion}{motion\xspace}
\newcommand{\motions}{motions\xspace}

\usepackage[pagebackref,breaklinks,colorlinks,allcolors=cvprblue]{hyperref}
\usepackage[capitalize]{cleveref}

\newcommand{\matheus}[1]{{\color{purple}Matheus: #1}}
\newcommand{\paul}[1]{{\color{orange}Paul: #1}}

%% file: sec/0_abstract.tex
\begin{abstract}

Predicting diverse object motions from a single static image remains challenging, as current video generation models often entangle object movement with camera motion and other scene changes. While recent methods can predict specific motions from motion arrow input, they rely on synthetic data and predefined motions, limiting their application to complex scenes. We introduce \methodname, a training-free approach that explores a pre-trained image-to-video generator’s latent distribution to discover various distinct and plausible motions focused on selected objects in static images. We achieve this by employing a flow generator guided by energy functions designed to disentangle object and camera motion.
Additionally, we use an energy inspired by particle guidance~\cite{particleGuidance24} to diversify the generated motions, 
without requiring explicit training data. Experimental results demonstrate that \methodname generates realistic and varied object animations, surpassing previous methods and even human predictions regarding plausibility and diversity. Code will be released upon acceptance. 
\end{abstract}

%% file: sec/1_intro.tex
\section{Introduction}
\label{sec:intro}

\setlength{\epigraphwidth}{0.9\linewidth}
\epigraph{Prediction is very difficult, especially if it's about the future.}{Niels Bohr}

\begin{figure}[t!]
    \centering
    \includegraphics[width=\columnwidth]{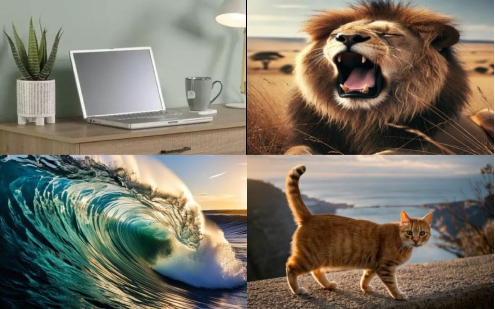}
    \caption{Could you imagine how the scene evolves in each case? See \cref{fig:teaser_part2} for plausible yet distinct motion videos predicted by our training-free approach \methodname.}
    \label{fig:teaser}
\end{figure}

Consider \cref{fig:teaser}. Can you imagine what could happen next in each case? Humans are good at imagining multiple ways the objects could move, even from single (image) snapshots. While we can train networks to predict videos starting from a conditioning text or image~\cite{blattmann2023SVD},
most generated videos
entangle camera motion, object motion, and other scene changes --
predicting a diverse set of motions for a given object still remains an open challenge.

Authoring plausible animations for objects in a static image can be daunting. Researchers have recently been able to train networks to predict cyclic and small-scale motions~\cite{li2024_GenerativeImageDynamics,bertiche2023blowingwindcyclenethuman}. Another family of  methods~\cite{li2024dragapart,shi2024motion} simplify this task by taking input motion arrows along with the starting image to predict videos with motions that follow the given arrows.
However, such methods
are trained on synthetic data and
do not generalize to complex motions, such as the breaking ocean wave in \cref{fig:teaser}. More importantly, they require motions to be given, rather than predicting them. In many scenarios, such as the roaring lion in \cref{fig:teaser}, imagining a diverse set of motions and then conveying them with multiple corresponding motion arrows itself, can be very challenging. 
%
The ability to automatically discover diverse yet plausible object motions can thus assist users in cinematic exploration, motion illustration, and image/video editing.

The latest image-to-video generators provide this opportunity. Having been trained on a large variety of diverse data, such generators, conditioned on static images, encode distributions over plausible animations for scene objects and other scene properties. Our paper subsequently asks and affirmatively answers the research question: \textit{is it possible to probe such a latent distribution to discover possible \motions for a given object in a static image}?

\begin{figure*}[t!]
    \centering
    \includegraphics[width=\linewidth]{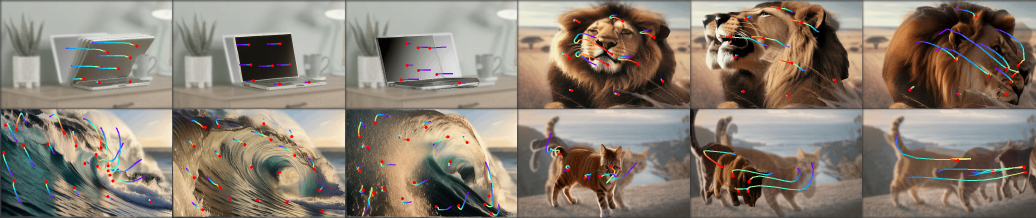}
    \caption{\textbf{\methodname} creates multiple distinct and plausible motions for a given object, disentangled from the motion of other objects, camera and other scene changes. We show three distinct object motions for each of Figure~\ref{fig:teaser}'s images, representative of constrained rigid motion (latop), complex deformations (wave) and articulated characters (lion and cat). 
    We visualize motions as flow trajectories from blue (first frame) to red (last frame). Ghosted intermediate frames further clarify complex motions. See supplemental for the result videos. }
    \label{fig:teaser_part2}
\end{figure*}

Directly sampling these generators, conditioned on a starting image, produces random videos, some of which may include a motion of the selected object. Still, most will consist of motions pertaining to other random objects, camera motion, lighting, and other changes to scene appearance.
Hence, the main challenges are to discover the \motions of an object in such a distribution that (i)~disentangle motions of the selected object from other
scene changes, and (ii)~find multiple distinct object motions. 
We propose \emph{\methodname} as a training-free method to find such object motions by exploring the prior of a pre-trained image-to-video generator.

We show that both of the above challenges can be addressed with a training-free approach that guides the denoising process of a flow generator~\cite{shi2024motion} with carefully designed guidance energies. Using a flow generator naturally disentangles objects and camera motion from other scene changes.
Our proposed guidance energies fulfill two purposes: (i)~they further disentangle object and camera motion by encouraging non-zero object motion and zero camera motion, and (ii)~encourage the generation of multiple distinct motions. We demonstrate that such guidance can be applied directly at inference time, without any fine-tuning of existing generators or access to suitable training data. \cref{fig:teaser_part2} shows the output of \methodname on the images shown in \cref{fig:teaser}. 


We evaluated \methodname on a variety of input images and compared ours with possible baselines (e.g., random sampling, LLM-based)
and ablated versions of our full method. We performed human evaluations to assess the quality of our generation, both in terms of plausibility and diversity of the predicted motions. The qualitative and quantitative evaluations show that we can reliably and accurately predict potential future outcomes, sometimes even surpassing human ability. We show that discovered \motions can be used for motion exploration and to facilitate drag-based image editing.
In summary, \methodname is the first training-free method to generate diverse and plausible videos of object motion from a single input image.

%% file: sec/2_related_work.tex
\section{Related Work}
\label{sec:related_work}

\noindent\textbf{Motion-aware video generators.}
Diffusion-based video generators have quickly advanced in the last years~\cite{ho2022video, blattmann2023stable, videoworldsimulators2024, polyak2024movie}, now producing realistic and temporally consistent videos.
Adding extra control, 
Motion-I2V~\cite{shi2024motion} introduced an image-guided video generation method as a two-stage process for consistent and controllable video generation. First, it uses a diffusion-based motion field predictor to determine pixel trajectories, followed by motion-augmented temporal attention that improves feature propagation across frames. We use this setup as our backbone and adapt it with our guidance energies during the denoising phase. 
AnimateAnything~\cite{dai2023animateanything} presents an image animation method using a video diffusion generator's motion prior, enabling controlled animation by guiding motion areas and speed. They demonstrate fine-grained, text-aligned animations with intricate motion sequences, even on open-world settings. Such methods, however, require suitable text prompts to guide the generation, which may be non-trivial in more complex scenarios where mentally predicting future \motions is challenging (see our LLM-based baseline in \cref{sec:results}). 
Finally, towards train-free methods, similar to the analysis of image generators~\cite{ganspace:2020}, Xiao et al.~\cite{xiao2024video} identify (using PCA analysis) motion-aware features in video 
diffusion models and
use them for interpretable and adaptable video motion control across different architectures.

\noindent\textbf{Generating cyclic motions.}
Creating future animations from static scenes has received attention over the years. Davis et al.~\cite{davis2015interactive} create interactive elements in videos by analyzing subtle object vibrations to get \motions, allowing manipulation of video elements as if they were physically interactive. 
The problem was recently revisited by Li et al.~\cite{li2024_GenerativeImageDynamics} to learn an image-space prior on scene motion from a collection of motion trajectories extracted from real video sequences depicting natural, oscillatory dynamics(e.g., leaves, trees, flowers, candles). Using a Fourier domain analysis, they learn a diffusion process to model the generation in the frequency space. 
Earlier, in the context of geometric objects, Mitra et al.~\cite{mitra_howThingsWork_sig_10} use symmetry analysis to infer plausible part movement in mechanical objects, focusing on gear assemblies and linkages. Hu et al.~\cite{partMobility:17} present a model for predicting part mobility in 3D objects by learning how parts of an object can move based on their spatial configuration in a single static snapshot by leveraging a linearity trait in typical object motions and creating a mapping that associates static snapshots with dynamic units.
To model small and repetitive garment motion, Bertiche et al.~\cite{bertiche2023blowingwindcyclenethuman} present an automatic method to generate human cinemagraphs from single RGB images to mimic garment dynamics arising from gentle winds.
They introduce a cyclic neural network that produces looping cinemagraphs for the target loop duration.
The network is trained with normal maps obtained from renderings of synthetic garment simulations.
While they demonstrated that the learned dynamics can be applied to real RGB images,
the reliance on training data does not allow these methods to be applied to the broader class of general motions.

\noindent\textbf{Movements from generative priors.} Priors learned by modern generators, trained on large datasets, have shown to be useful for handle-based image manipulation. 
DragGAN~\cite{dragGAN23} presented an interactive tool for handle-based realistic editing of natural images that relied on a feature-based motion supervision that moves selected points toward target positions, leveraging GAN's internal features for precise localization. Similarly, image manipulators (e.g., point- or box-based) have exploited priors implicit in diffusion-based image~\cite{shi2023dragdiffusion,avrahami2024diffuhaul,pandey2024diffusionhandles,mou2024dragondiffusion} or video~\cite{shi2024lightningdrag,xiao2024video,wang2024boximator} generators. 
Beyond zeroshot methods, Dragapart~\cite{li2024dragapart} presents a part-level editing system where they refine a pretrained image generator on a new synthetic dataset showing annotated part motion. The network, fine-tuned on synthetic data generalizes well across real-world images and diverse categories. However, the method fails on complex scenarios and object categories not seen in the training set (\cref{sec:results}). Draganything~\cite{wu2024draganythingmotioncontrolusing} uses entity representation for drag-based plausible video generation in response to user arrows, but does not produce diverse results.
There are also sampling strategies designed to increase the diversity of outputs in diffusion-based image generators~\cite{particleGuidance24}.
They rely on concurrently denoising a batch of multiple samples guided by a repulsive energy.
However, in the case of video generators, such strategies are limited by the memory cost of the number of samples that can be denoised together ($\approx$10 GB per additional sample for Motion-I2V \cite{shi2024motion} with gradient checkpointing).
On the other hand, we devise an iterative sampling strategy that is not capped by the number of samples that can fit in memory together (see Section~\ref{sec:stopping}).


%% file: sec/3_method.tex
\section{Method}
\label{sec:method}

\newcommand{\x}{\mathbf x}
\newcommand{\X}{\mathcal{X}}
\newcommand{\y}{\mathbf y}
\newcommand{\m}{\mathbf m}

\begin{figure}
    \centering
    \includegraphics[width=\linewidth]{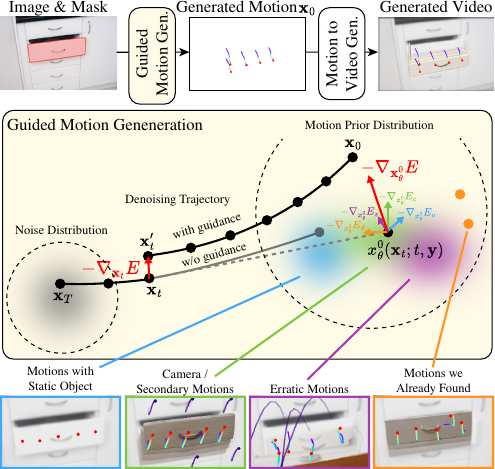}
    \caption{\textbf{Method Overview.} We generate a motion $\x$ using a guided denoising approach, where guidance energies encourage smooth object motions that are disentangled from camera motions and distinct from previously generated motions. Iterative sampling gives us a set of diverse motions $\X$.}
    \label{fig:overview}
\end{figure}

Our goal is to take an image $\y \in \mathbb{R}^{H \times W \times 3}$ and a mask $\m \in \mathbb{R}^{H \times W}$ marking an object in the image, and to find a set of likely \emph{\motions} $\X \coloneq \{\x^{(1)}, \x^{(2)}, \dots \}$ of the object given its context in the image. In Figure~\ref{fig:overview}, for example,
the drawer could be opened or closed; however it could not plausibly be moved up- or downwards.
We represent motions as time-dependent two-dimensional vector fields $\x \in \mathbb{R}^{F \times H \times W \times 2}$ for a motion that spans $F$ frames. This vector field defines the trajectory of each pixel as per-frame 2D offsets from its initial position.

We generate \motions by sampling an existing image-to-video diffusion model that takes the image $\y$ as starting frame. The main challenges for generating \motions of an object in an image are disentangling object motions from other types of scene changes and finding a diverse set of plausible object motions. To address these challenges, we
(i)~use a diffusion model that generates motion separately from appearance~\cite{shi2024motion}, effectively disentangling object / camera motions from other scene changes, such as lighting or shadows (Section~\ref{sec:generation}), and
(ii)~define guidance energies that we minimize during the denoising process to further separate object motion from camera motion and to efficiently sample a diverse set of motions, rather than sampling motions randomly from the motion prior (Section~\ref{sec:guidance}).
We build the motion set $\X$ by iteratively sampling the motion prior with our guidance energies, and define a simple stopping criterion to avoid implausible motions (Section~\ref{sec:stopping}).

\subsection{Motion Generation}
\label{sec:generation}

Our approach can be applied to any pre-trained diffusion-based image-to-video model which generates motion and appearance independently. 

\noindent\textbf{Training.}
Given an input image $\y$ and a motion $\x$ for this image,
a noisy motion vector field $\x_t$ is first obtained by adding a random amount of noise to $\x$:
\begin{equation}
    \label{eq:noising}
    \x_t = \sqrt{\alpha(t)}\ \x + \sqrt{1-\alpha(t)}\ \epsilon,
\end{equation}
where $\epsilon \sim \mathcal{N}(\mathbf{0}, \mathbf{I})$ is Gaussian noise, and $t \in [0, T]$ parameterizes a noise schedule $\alpha$ that determines the amount of noise in $\x_t$, with $\alpha(0) = 1$ (no noise) and $\alpha(T) = 0$ (pure noise).
The denoiser $\epsilon_{\theta}$ of the diffusion model is then trained
to invert this noising process by minimizing the following loss through gradient-descent: 
\begin{equation*}
    \label{eq:diffloss}
    \mathcal{L_{\text{diff}}} := w(t) \|\epsilon_\theta(\x_t; t, \y) - \epsilon\|_2^2,
\end{equation*}
where
$w(t)$ is a weighting scheme for different parameters $t$.
In practice, we employ a latent-space diffusion model that operates on a lower-resolution latent representation of the motions and the input image, which is obtained through a VAE~\cite{kingma2013auto}. We omit this distinction in the notation, both for clarity and for generality, as the method is orthogonal to the choice of diffusion model. Specifically, our implementation utilizes Motion-I2V~\cite{shi2024motion} as the backbone. 


\noindent\textbf{Inference.}
Given the trained denoiser $\epsilon_\theta$, a noise-free motion $\x_0$ for input image $\y$ is then generated by starting from pure noise $\x_T$ and iteratively denoising in small steps:
\begin{align}
    \x_T &\sim \mathcal N(\mathbf 0, \mathbf I) \nonumber \\
    \x_{t-1} &\sim \mathcal{N}(a_t \x_t - b_t \epsilon_\theta(\x_t; t, \y), \sigma^2_t\mathbf{I})
    \label{eq:inference_step}
\end{align}
where $a_t$, $b_t$, and the variance $\sigma^2_t$ are chosen according to a denoising schedule. This process creates a trajectory 
$\x_T, \x_{T-1}, ..., \x_0$ of motions with decreasing noise, where $\x_0$ is close to the natural motion manifold. Generated motions $\x_0$ are typically plausible, but they entangle camera motions with object motions. Additionally, exploring different \motions by randomly sampling $\x_T$ is inefficient, as it requires a large number of samples to find multiple meaningful ways in which $\y$ can change in time.

\subsection{Guidance Energies}
\label{sec:guidance}

Our key contribution is the guidance energies that we introduce into the inference process. The energies encourage the generation of motions that are different from any previously generated motions, where only the object in image $\y$ selected by the mask $\m$ moves and the camera is static. The goal is to significantly reduce the number of samples needed to get a diverse set of focused object motions.

\noindent\textbf{(i)~Static camera guidance.}
We encourage zero camera motion by penalizing the average magnitude of motion outside the object region defined by the object mask $\m$:
\begin{equation*}
    \label{eq:camera_guidance}
    E_c(\x,\m) \coloneq \frac{\sum_{k,i,j} \|\x_{k,i,j}\|\ (1-\m_{i,j})}{\sum_{k,i,j} (1-\m_{i,j})},
\end{equation*}
where $k, i, j$ are indices over frames, pixel rows, and pixel columns, respectively, so that $\x_{k,i,j}$ denotes a single offset vector of the motion $\x$. The mask $\m$ is $1$ inside the object region and $0$ everywhere else.

\noindent\textbf{(ii)~Object motion guidance.}
We encourage object motion by encouraging a difference between the average magnitude of motion inside the object mask $\m$ and outside:
\begin{equation*}
\label{eq:object_guidance}
    E_o(\x,\m) \coloneq \phi\left(\left|E_c(\x,\m) - E_c(\x,1-\m)\right|\right).
\end{equation*}
Here, $\phi$ is an activation function that gives higher energies for smaller differences, based on a soft inverse:
\begin{equation*}
    \phi(a) \coloneq \text{softplus}\left((a + e)^{-1} - \tau\right),
\end{equation*}
where $e$ is a small epsilon to avoid division by zero, and $\tau$ is a threshold representing the point at which a satisfactory loss value is reached. $\tau$ is empirically set to $40$ for the object motion guidance and $1$ for the diversity guidance.

\noindent\textbf{(iii)~Diversity guidance.}
Given a set of previously generated motions $\X$, we encourage newly generated motions to be different by adding a repulsion energy from each of the motions in $\X$:
\begin{equation*}
    \label{eq:diversity_guidance}
    E_d(\x,\m, \X) \coloneq \sum_{\tilde{\x} \in \X} \frac{\sum_{k,i,j} \phi\left(d(\x_{k,i,j},\tilde{\x}_{k,i,j})\right)\ \m_{i,j}}{\sum_{k,i,j} \m_{i,j}},
\end{equation*}
where $d$ is a distance function between individual offset vectors in a motion based on angle and magnitude differences:
%
\begin{equation*} 
d(\mathbf{a}, \mathbf{b}) \coloneq w_{\text{mag}} (\left|\|\mathbf{a}\|  - |\mathbf{b}\|\right|) + w_{\text{angle}}\left(1 - \frac{\mathbf{a}^\top \mathbf{b}}{\|\mathbf{a}\|\ \|\mathbf{b}\|}\right),
\end{equation*}
with weights $w_{\text{mag}}=0.25$ and $w_{\text{angle}}=0.75$ to emphasize diverse motion directions. 


\noindent\textbf{(iv)~Smoothness guidance.}
As a regularization, we also encourage smooth object motions by penalizing large changes in motion across consecutive frames within the object mask:
\begin{equation} \label{eq: smoothness_guidance} 
E_s(\x, \m) \coloneq \frac{ \sum_{k,i,j} d\left( \x_{k,i,j}, \x_{k+1,i,j} \right) \m_{i,j}}{\sum_{k,i,j} \m_{i,j}}, 
\end{equation}
with $w_{\text{mag}}=0.75$ and $w_{\text{angle}}=0.25$ to minimize sudden changes in magnitude.

\noindent\textbf{Guided Inference.} We combine the four energies into a single (guidance) energy $E(\x,\m,\X) \coloneq \lambda_d E_d + \lambda_c E_c + \lambda_o E_o + \lambda_s E_s,$ with weights $\lambda_d = 3.0$, $\lambda_c = 0.2$, $\lambda_o = 0.025$ and $\lambda_s = 0.1$. Similiar to classifier-free guidance and several image editing methods~\cite{ho2022classifierfreediffusionguidance, epstein2023diffusionselfguidancecontrollableimage, pandey2024diffusionhandles}, we minimize these energies during the inference process, effectively changing the denoising trajectory, without requiring fine-tuning or retraining (which would be difficult as our tasks lacks suitable training data). Equation~\ref{eq:inference_step} takes the modified form:
\begin{align}
    \x_{t-1} &\sim \mathcal{N}\left(a_t \x_t - b_t \epsilon_\theta(\x_t^\prime; t, \y\right), \sigma^2_t\mathbf{I}) \text{, with} \\
    \x_t^\prime &\coloneq \x_t - \nabla_{\x_t} E\left(x_\theta^0(\x_t; t, \y), \m, \X\right). \nonumber
\end{align}
Here, $x_\theta^0(\x_t; t, \y)$ is the non-noisy motion predicted at inference step $t$, derived from Eq.~\ref{eq:noising} as:
\begin{equation*} 
    x_\theta^0(\x_t; t, \y) \coloneq \frac{1}{\sqrt{\alpha(t)}} \left(\x_t - \sqrt{1-\alpha(t)}\ \epsilon_\theta(\x_t; t, \y)\right).
\end{equation*}

\subsection{Stopping Criterion}
\label{sec:stopping}
We build the set $\X \coloneq \{\x^{(1)}, \x^{(2)}, \dots \}$ by iteratively sampling the motion prior as described above. We can obtain an arbitrary number of motions $\x$ using this strategy; for our experiments we sample up to $6$ different \motions. However, some objects and scenes may only admit a smaller number of distinct motions, after which motions either repeat or stop. We detect these cases using the guidance energy of the final denoised motion $E(\x_0, \m, \X)$. We discard and re-sample \motions with guidance energies above a threshold $\rho ( = 5.0)$, and stop sampling after discarding two \motions in a row. 








%% file: sec/4_results.tex
\section{Results}
\label{sec:results}

\begin{figure*}[t]
    \centering
    \includegraphics[width=\linewidth]{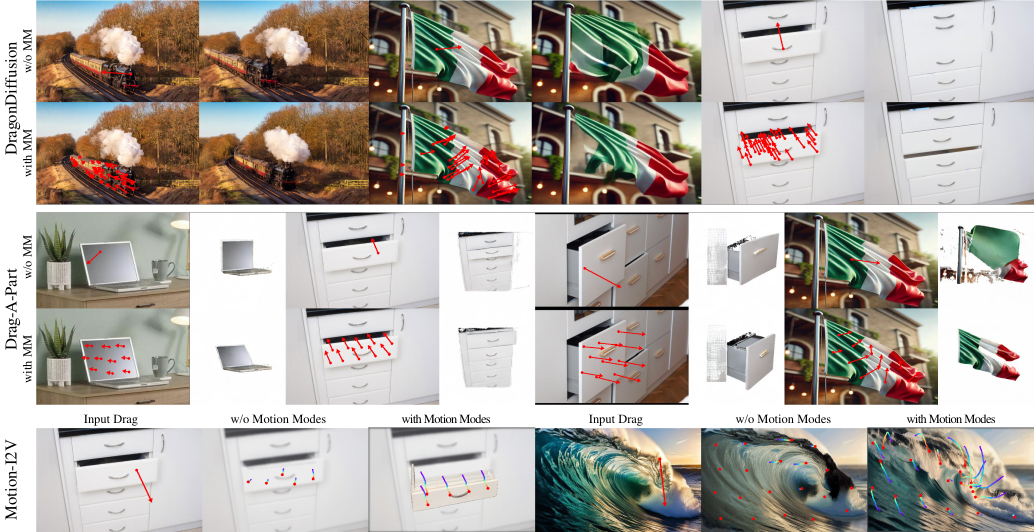}
    \caption{\textbf{Motion Completion.} We can use our set of \motions $\X$ to complete rough motion hints (single red arrows) given by the user as conditional input to either drag-based image editors like DragonDiffusion or Drag-A-Part, or motion-to-video generators like Motion-I2V. Using the more detailed \motions allows for complex motions that are hard to specify manually (like the flag or wave animations), and avoids ambiguities in the conditional input that can lead to implausible results, like the floating train, or the squashed drawer.}
    \label{fig:snapping_application}
    \vspace{-10pt}
\end{figure*}

To evaluate a set of \motions $\mathcal{X}$ generated by our method, we identify four desirable motion properties:
(1) \emph{Plausible}: motions appear natural and physically reasonable. (2) \emph{Diverse}: motions are largely different from each other. (3) \emph{Expected}: motions are plausible motions that match those imagined by a viewer for the selected object in the image.
(4) \emph{Focused}: motions avoid any scene motion (including camera motion) that does not pertain to the selected object or is directly caused by its motion (eg. the smoke from the selected train (top-left) in Figure~\ref{fig:snapping_application}).


We show with both quantitative metrics and a user study that our guided sampling strategy performs significantly better along these properties than alternatives, given the same sample budget and the same motion prior.
We also provide several qualitative comparisons that demonstrate that \methodname can be used to explore object motions.
As additional application, we also show how our motions can be used to assist users with drag-based image editing.
More evaluation results are provided in the supplement.

\noindent\textbf{Baselines.}
As far as we know, \methodname is the first training-free method to explore the problem of finding diverse motions for a given object in an image. However, there are several alternatives we can compare against. For a fair comparison, all baselines use the same Motion-I2V~\cite{shi2024motion} backbone as our method.
(1) \emph{Prompt Generation}: We give GPT4-o an image with highlighted object and ask it to give us prompts for diverse object motions, which we then feed into Motion-I2V. Each prompt gives us one \motion.
(2) \emph{ControlNet}: We use Motion-I2V's \emph{MotionBrush} to restrict motions to the object region. This tool is a ControlNet trained to limit motions to originate in the given region. We obtain multiple \motions by randomly sampling the starting noise $\x_T$.
(3) \emph{Random Arrows}: We use Motion-I2V's \emph{MotionDrag} with random arrows to explore possible object motions. We sample an arrow by choosing a random starting position inside the object region, a random direction and a fixed length. Each arrow gives us a different \motion.
(4) \emph{Random Noise}: We randomly sample the starting noise $\x_T$ of Motion-I2V. This is equivalent to our method without any guidance energies.
(5) \emph{Farthest Point Sampled (FPS) Noise}: We use farthest point sampling to sample distinct starting noise $\x_T$.

\begin{figure*}[t]
    \centering
    \includegraphics[width=\linewidth]{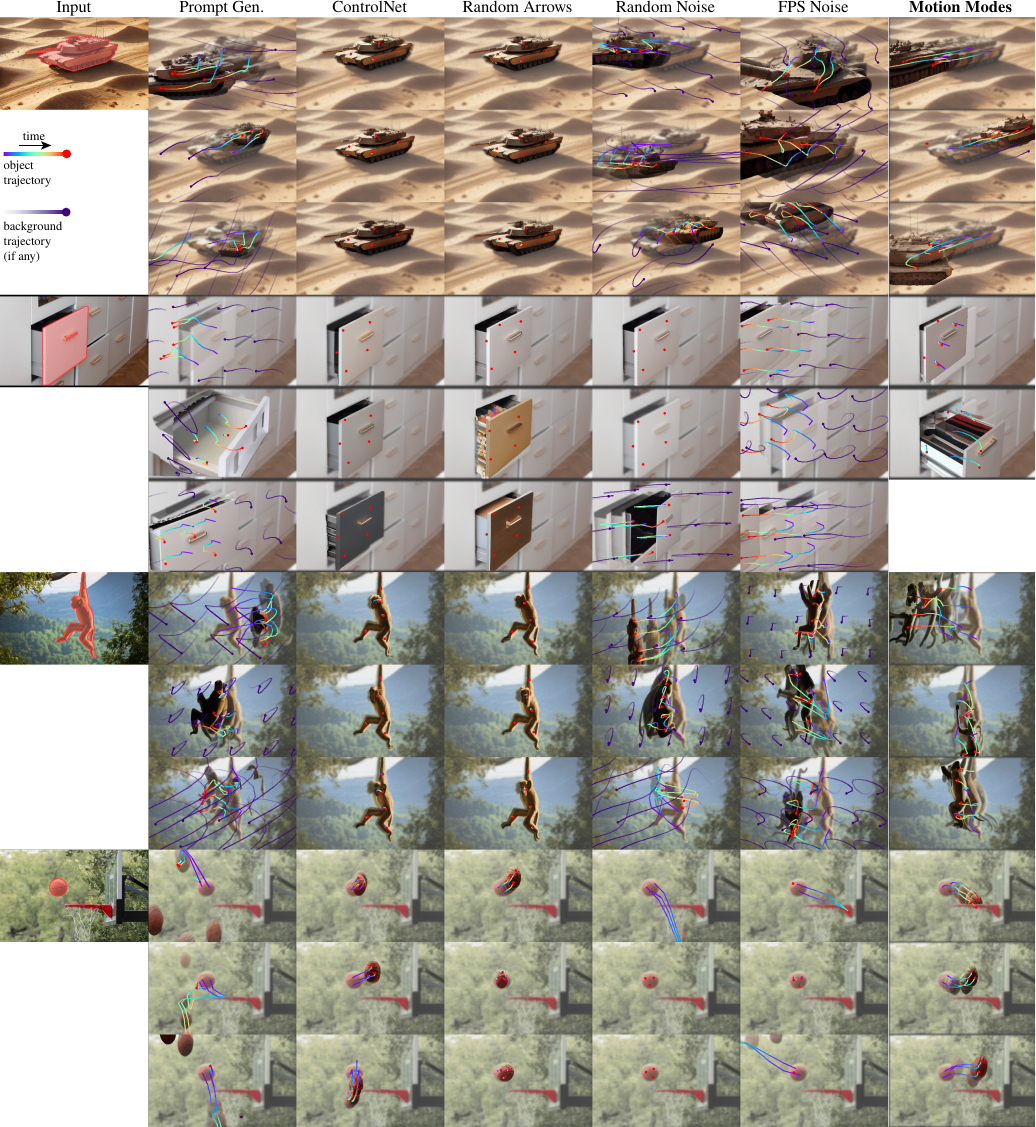}
    \caption{\textbf{Qualitative comparison.} Each column shows the first three \motions for the masked object in the input (left). Object trajectories have red endpoints, background trajectories (usually due to camera motion) are purple. Motion is additionally visualized by overlaying ghosted intermediate frames.
    We can see that \methodname finds more plausible and diverse object motions disentangled from any other motions or scene changes, such as camera motions.}
    \label{fig:qualitative_comparison}
    \vspace{-10pt}
\end{figure*}

\paragraph{Qualitative comparison.}
Figure~\ref{fig:qualitative_comparison} shows a qualitative comparison to all baselines on four scenes. (Please refer to the supplement for a comparison on a larger set of images.) We can see that the prompt generation baseline does tend to generate motions that are \emph{diverse}, but the inaccurate nature of the prompt-based control results in less \emph{focused} motions of the selected object. There is significant camera motion, and we can see motions of secondary objects in the basketball image, for example, where additional balls are hallucinated. Restricting the motion to the object region using the ControlNet baseline has the undesirable effect of significantly reducing the overall amount of motion, to the point of resulting in completely static scenes in many cases. Similar to the prompt generation baseline, sampling the motion prior randomly or with farthest point sampling without using our guidance energies entangles object motion with camera motion.
Additionally, we can see that our approach produces more \emph{plausible} and \emph{expected} motions,
compared to all baselines.
For example, the opening and closing motion of the drawer is more natural without deforming parts, and the forward/backward motion of the tank generated by \methodname is probably closer to the motion we would expect from the tank than the more erratic motions generated by the baselines. We further confirm this trend on a larger set of scenes with the user study presented in the one of the following sections. We attribute the improved plausibility to our smoothness energy that avoids erratic motions.


\paragraph{Quantitative comparison.}
\begin{table}[]
    \centering
    \caption{
    \textbf{Quantitative comparison} of the \emph{diverse} and \emph{focused} property of our output motions to all baselines.
    }
    \footnotesize 
    \renewcommand{\arraystretch}{1.0}
    \setlength{\tabcolsep}{5pt}
    \begin{tabularx}{\linewidth}{r >{\centering\arraybackslash}X >{\centering\arraybackslash}X @{\extracolsep{\fill}} >{\centering\arraybackslash}X @{\extracolsep{\fill}} >{\centering\arraybackslash}X} 
    \toprule
         

         & \texttt{diverse} & \multicolumn{3}{c}{\texttt{focused}} \\
         \cmidrule(l    r){2-2} \cmidrule(l){3-5}
         & $\bar{E}_d\downarrow$ & $\bar{E}_f\downarrow$ & ($\bar{E}_c\downarrow$ & $\bar{E}_o\downarrow$) \\
         
         
         \midrule
Prompt Gen. & 1.28 & 1.71 & 1.11 & 2.31 \\
ControlNet & 1.75 & 1.14 & \textbf{0.07} & 2.22 \\
Random Arrows & 1.77 & 1.17 & \textbf{0.07} & 2.27 \\
Random Noise & 1.27 & 2.20 & 1.36 & 3.05 \\
FPS Noise & 1.21 & 1.98 & 1.23 & 2.74 \\
\midrule
\textbf{\methodname} (ours) & \textbf{1.04} & \textbf{0.07} & \textbf{0.09} & \textbf{0.05} \\
         \bottomrule
    \end{tabularx}
    \label{tab:quantitative_comparison}
\end{table}
We measure two
properties with explicit quantitative metrics:
First, the \emph{diversity} of motions in a set $\X$ can be measured with the average diversity guidance energy $\bar{E}_d(\m, \X) \coloneq \sum_{\x \in \X} E_d(\x, \m, \X) / {|\X|} $.
Second, the \emph{focus} of motions on only the selected object can be measured based on the average object motion and static camera guidance energies $\bar{E}_f \coloneq 0.5 (\bar{E}_o + \bar{E}_c$), with $\bar{E}_o$ and $\bar{E}_c$ computed analogous to $\bar{E}_d$, but scaled by a factor of $0.01$ and $0.1$, respectively, to account for scaling differences.

We compare our method to all baselines on a dataset of $28$ input images that were obtained either through a state-of-the-art text-to-image generator, or from photographs. The images cover a wide range of scenes, including articulated objects, vehicles, animals, balls, and objects with and objects with complex motions, such as waves and flags. Please refer to the supplement for a full set of qualitative results.

Table~\ref{tab:quantitative_comparison} summarizes results. Due to our diversity guidance, we achieve significantly more diverse motions than any baseline. The ControlNet and Random Arrows baselines also achieve relatively good focus, but looking at Fig.~\ref{fig:qualitative_comparison} (as well as the camera and object guidance columns in Table~\ref{tab:quantitative_comparison}), we can see that this is mostly caused by a lack of both camera \emph{and} object motion. Our guidance energies fix the camera without fixing the object, giving us more focused motions.

        
\paragraph{User studies.}
We perform two user studies. The first study evaluates the \emph{plausible}, \emph{diverse} and \emph{expected} nature of our output motions compared to baselines, while the second study examines the \emph{expected} nature of our motions. 

\begin{figure}[t]
    \centering
    \includegraphics[width=\linewidth]{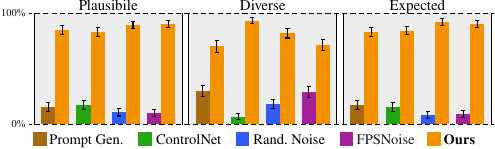}
    \caption{\textbf{User Study I.} We compare the \emph{plausible}, \emph{diverse}, and \emph{expected} nature of our motions to four baselines. Each pair of bars shows the percentage of comparisons in which our method or a baseline was judged favorably with $95$\% confidence intervals.}
    \label{fig:user_study}
\end{figure}

In the first study, participants were asked to compare the top three \motions of our method to top three \motions of a baseline, and choose the best set of \motions along each of the three metrics in three two-alternative forced choice questions. The methods were presented in a randomized order. We recruited $32$ participants, each completed $10$ comparisons per baseline (a total of $320$ comparisons per baseline). For each comparison, a scene was chosen randomly from our dataset of $27$ images. Results are shown in Figure~\ref{fig:user_study}: motions of our approach are judged to be more plausible, diverse and expected than motions found by baselines.
Notedly, the prompt generation baseline also has a good amount of diversity, coming close to the diversity of our approach.
%
We omitted the \emph{Random Arrows} baseline due to its similarity (and worse performance) compared to ControlNet. It is included in an extended version of the study in the supplement.

In the second study, 12 new participants were first asked to describe all possible future motions of an object highlighted in an input image. We then revealed the first four of our \motions to them, and asked them to make two  independent sets of selections - (i)~\motions that align with their initial expectations and (ii)~\motions that are plausible. Each participant assessed 10 scenes, and
we computed three metrics from their responses - \emph{expected} (percentage of their expected motions predicted by our \motions), \emph{plausible} (percentage of our \motions deemed plausible), and \emph{inspirational} (percentage of our \motions that were deemed plausible but outside the participant’s expectation). Participants found on average, that $96$\% of \motions were plausible, $92$\% of their expectations were produced by our approach, and $19$\% of \motions were plausible but outside expectation. Overall, participants felt that our \motions not only aligned well with their expectations, but also consistently provided inspiration for exploring unseen diverse motions in input scenes.

\paragraph{Ablation.}
\begin{table}[]
    \centering
    \caption{
    \textbf{Ablation} of key components with metrics based on \emph{diverse}, \emph{focused} metrics and their tradeoff $\bar{E} \coloneq 0.5 (\bar{E}_d + \bar{E}_f)$. Underlined values are closer to the best than to the worst value.
    }
    \footnotesize 
    \renewcommand{\arraystretch}{1.0}
    \setlength{\tabcolsep}{5pt}
    \begin{tabularx}{\linewidth}{r >{\centering\arraybackslash}X >{\centering\arraybackslash}X >{\centering\arraybackslash}X @{\extracolsep{\fill}}  > {\centering\arraybackslash}X @{\extracolsep{\fill}}  >{\centering\arraybackslash}X} 
    \toprule
         
         & & \texttt{div.} & \multicolumn{3}{c}{\texttt{focused}} \\
         \cmidrule(l    r){3-3} \cmidrule(l){4-6}
         & $\bar{E}\downarrow$ & $\bar{E}_d\downarrow$ & $\bar{E}_f\downarrow$ & ($\bar{E}_c\downarrow$ & $\bar{E}_o\downarrow$) \\
         
         
         \midrule
without $E_c$ & 0.83 & \underline{1.02} & 0.64 & 1.29 & \textbf{0.00} \\
without $E_o$ & 0.97 & \underline{1.03} & 0.91 & \textbf{0.06} & 1.75 \\
without $E_d$ & 0.72 & 1.36 & \underline{0.08} & \underline{0.13} & \underline{0.04} \\
FPS instead of $E_d$ & 0.79 & 1.49 & \underline{0.10} & \underline{0.11} & \underline{0.08} \\
ControlNet instead of $E_c$,$E_o$ & 0.88 & \textbf{0.96} & 0.80 & \underline{0.15}
 & 1.45 \\
\midrule
\textbf{\methodname} & \textbf{0.55} & \underline{1.04} & \textbf{0.07} & \underline{0.09} & \underline{0.05} \\
         \bottomrule
    \end{tabularx}
    \label{tab:ablation}
\end{table}

We ablate several components of our method is shown in Table~\ref{tab:ablation}. We use the same metrics as in Table~\ref{tab:quantitative_comparison}, but add another metric that illustrates the trade-off between diversity and focus: $\bar{E} \coloneq 0.5 (\bar{E}_d + \bar{E}_f)$. We ablate the three main guidance energies, and show the effect of using farthest point sampling of the initial noise instead of the diversity guidance, and a ControlNet instead of the camera and object guidance. As expected, removing the camera or object guidance results in strong camera motions (high $\bar{E}_c$) or little object motion (high $\bar{E}_o$), and removing the diversity energy or using farthest point sampling instead results in less diversity (high $\bar{E}_d$). Swapping the object and camera guidance with a ControlNet tends to fix the object in place (high $\bar{E}_o$). We only achieve the best tradeoff between diversity and focus using all of our components.

\paragraph{Application.} 
\methodname, as presented, can help artists efficiently explore a diverse set of motions for a selected object, without having to sieve through a large set of sampled  videos containing disentangled object and camera motion.

\textit{Arrow-based motion prompting.} 
We demonstrate a second application: completing a rough motion hint to be used as input to a drag-based image editor or a motion-to-video generator. Figure~\ref{fig:snapping_application} shows examples on two recent drag-based image editors: DragonDiffusion~\cite{mou2024dragondiffusion} and Drag-A-Part~\cite{li2024dragapart}, and one motion-to-video generator~\cite{shi2024motion}, comparing results with and without our motion completion.
A single drag arrow given by the user (shown in red) is used to retrieve the closest one of our detailed \motions (shown as multiple red arrows for the drag-based image editors). We define the closest \motion as containing the 2D offset with lowest distance to the provided drag arrow, across all frames of the \motion. We then use this \motion, instead of the original drag arrow, as conditional input to image editor or video generator. Please refer to the supplement for details.
%
%
This has two benefits: (i)~Specifying complex image edits or video motions in detail is both difficult and time consuming, thus obtaining a complex edit/motion from a quick hint saves time and does not require artistic expertise. For example, it would be difficult to manually construct detailed drag arrows for the flag or the ocean wave. (ii)~Rough motion hints are ambiguous and may be misinterpreted by the conditional generators, resulting in implausible motions. For example, dragging the train backwards with Dragon Diffusion results in a floating train, or dragging the drawer towards its closed position is misinterpreted as moving it upwards. Providing a more detailed motion removes this ambiguity and avoids implausible results.



%% file: sec/5_discussion.tex
\section{Conclusion}
\label{sec:discussion}

We have presented \methodname as a training-free method to discover distinct motions for a selected object mask in a static image. 
Our primary contribution is a novel combination of guidance energies applied at inference time, to sample videos showing diverse object motions, from a pre-trained diffusion-based video generator.  We evaluated our method on a range of complex images with both animate and inanimate objects to discover non-trivial \motions, sometimes beyond those anticipated by viewers.

\begin{figure}[h!]
    \centering
    \includegraphics[width=\linewidth ]{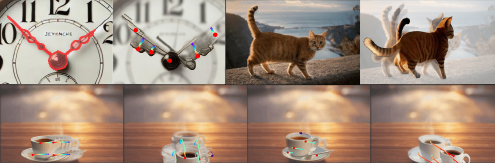} \caption{\textbf{Limitations.} (Top) The video prior can limit quality (bent clock handles, two cat tails). (Bottom) Continuous motion spaces can only be sampled discretely.}
\vspace*{-.15in}   
    \label{fig:limitations}
\end{figure}

\noindent\textbf{Limitations.} 
\cref{fig:limitations} shows example limitations. 
Foremost, since \methodname is training-free, we inherit any data bias in our video generator (e.g., we will miss motions that cannot be expressed in our generator's sampling space). 
As we currently seek a discrete set of \motions, we are only able to represent a continuous subspace of plausible motions a distinct set of discrete motions (eg. the laptop moving left-right, and front-back, instead of anywhere on the desk in \cref{fig:teaser_part2}).
Further, since we progressively generate \motions, we need a number of forward passes equal to the number of extracted \motions. This can be slow and undesirable. 
Finally, very specific underlying modes can produce unrealistic motions.

\noindent\textbf{Future work.}
\methodname produces videos with negligible camera motion. Extending our approach to generate object motion with moving cameras, as commonly observed in sport and action shots where the camera follows the trajectory of the moving object, is subject to future work.  We would also like to extend our method beyond 2D motion fields, to
produce 3D \motions: this would allow us to 
directly output 4D dynamic shapes as animated mesh sequences, turning video generators into 4D asset generators.

%% file: sec/X_suppl.tex
\section{Overview}
\label{sec:supp_overview}
In this appendix, we present extended versions of the user study (Section~\ref{sec:supp_user_study}) and the ablation study (Section~\ref{sec:supp_ablation}). Additionally, we examine how much a given motion constrains the video generator by showing different videos generated for the same motion (Section~\ref{sec:videos_for_one_motion}) and provide additional implementation details as well as timing details (Section~\ref{sec:timing}). Finally, we provide a more detailed description for some of the baselines (Section ~\ref{sec:baseline_details}) and the arrow-based motion prompting application (Section ~\ref{sec:snapping_details}).

Our project website, \url{https://motionmodes.github.io}, also contains, among other details, a full qualitative comparison on $28$ images, results of our method on a total of $34$ different input images, and our arrow-based motion prompting application using a different video generator~\cite{niu2024mofa}.

\section{Extended User Study}
\label{sec:supp_user_study}

\begin{figure}[t]
    \centering
    \includegraphics[width=\linewidth]{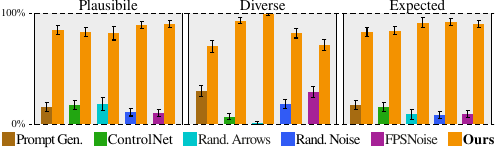}
    \caption{\textbf{Extended user study.} We compare the \emph{plausible}, \emph{diverse}, and \emph{expected} nature of our motions to five baselines, including the Random Arrows baseline. Each pair of bars shows the percentage of comparisons in which our method or a baseline was judged favorably with $95$\% confidence intervals.}
    \label{fig:user_study_extended}
\end{figure}

In Figure~\ref{fig:user_study_extended}, we present an extended version of the user study that includes the random arrows baseline. Results for this baseline are collected from $16$ instead of $32$  participants, the other study details are the same as for all other baselines. Results confirm our findings for all other baselines: users find our motions significantly more plausible and diverse, and they also better agree with the motions users expected for the selected object.

\section{Extended Ablation}
\label{sec:supp_ablation}

\begin{table}[]
    \centering
    \caption{
    \textbf{Extended ablation} of key components with metrics based on \emph{diverse}, \emph{focused} metrics and their tradeoff $\bar{E} \coloneq 0.5 (\bar{E}_d + \bar{E}_f)$. Underlined values are closer to the best than to the worst value.
    }
    \footnotesize 
    \renewcommand{\arraystretch}{1.0}
    \setlength{\tabcolsep}{5pt}
    \begin{tabularx}{\linewidth}{r >{\centering\arraybackslash}X >{\centering\arraybackslash}X >{\centering\arraybackslash}X @{\extracolsep{\fill}}  > {\centering\arraybackslash}X @{\extracolsep{\fill}}  >{\centering\arraybackslash}X} 
    \toprule
         
         & & \texttt{div.} & \multicolumn{3}{c}{\texttt{focused}} \\
         \cmidrule(l    r){3-3} \cmidrule(l){4-6}
         & $\bar{E}\downarrow$ & $\bar{E}_d\downarrow$ & $\bar{E}_f\downarrow$ & ($\bar{E}_c\downarrow$ & $\bar{E}_o\downarrow$) \\
         
         
         \midrule
without $E_c$ & 0.83 & \underline{1.02} & 0.64 & 1.29 & \textbf{0.00} \\
without $E_o$ & 0.97 & \underline{1.03} & 0.91 & \textbf{0.06} & 1.75 \\
without $E_d$ & 0.72 & 1.36 & \underline{0.08} & \underline{0.13} & \underline{0.04} \\
without $E_s$ & \underline{0.58} & \underline{1.02} & \underline{0.13} & \underline{0.10} & \underline{0.16} \\
FPS instead of $E_d$ & 0.79 & 1.49 & \underline{0.10} & \underline{0.11} & \underline{0.08} \\
ControlNet instead of $E_c$,$E_o$ & 0.88 & \textbf{0.96} & 0.80 & \underline{0.15}
 & 1.45 \\
\midrule
\textbf{\methodname} & \textbf{0.55} & \underline{1.04} & \textbf{0.07} & \underline{0.09} & \underline{0.05} \\
         \bottomrule
    \end{tabularx}
    \label{tab:ablation_extended}
\end{table}

In Table~\ref{tab:ablation_extended}, we provide an extended ablation study that includes an ablation of the smoothness guidance. Apart from its function as regularizer, surprisingly, this energy also improves object focus, i.e. it tends to better avoid static objects. Our interpretation is that object motions are suppressed by the motion generator's prior during the denoising process if they start out unrealistically jerky or jittery. Our smoothness energy guides the denoising trajectory away from these bad object motions early on, resulting in a less suppression from the prior.

\section{Multiple Videos Generated for One Motion}
\label{sec:videos_for_one_motion}
\begin{figure}[t]
    \centering
    \includegraphics[width=\linewidth]{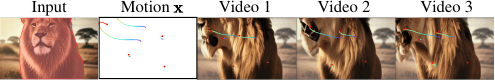}
    \caption{\textbf{Multiple videos from one motion.} We generate multiple videos from the same motion $\x$. They differ in small details, but overall follow the motion accurately.}
    \label{fig:multi_video_same_motion}
\end{figure}
All videos in our experiments are obtained by first generating a motion $\x$ and then generating a video conditioned on $\x$. To examine how closely the generated video follows $\x$, in Figure~\ref{fig:multi_video_same_motion}, we show multiple videos generated conditioned on the same motion $\x$ from different random noises. We can see that small details are different, but overall, the motions of the different videos are similar to each other and follow the generated motion $\x$ accurately.

\section{Implementation Details}
\label{sec:timing}

\paragraph{Guided Denoising}
As described in the paper, we use the flow generation module from Motion-I2V~\cite{shi2024motion} as our backbone. We further disconnect the ControlNet module described in their paper, as we don't need the conditioning and we found that the constraints from ControlNet conditioning limits the diversity of our motions. The flow generator uses 25 total timesteps for denoising out of which the first 20 timesteps are guided in our approach. 
\paragraph{Timing and Memory}
In our experiments, we further used gradient checkpointing on the U-Net to minimize the memory cost of backpropagating the guidance gradients in each denoising timestep. Given the time cost of gradient checkpointing and additional memory costs of backpropagation, our guided denoising approach has a peak memory usage of 21.7GB and requires on average 2 minutes 35 seconds to fully denoise a sample across 25 timesteps. Unguided vanilla denoising, on the other hand, has 12.3GB peak memory usage and requires 1 minute 18 seconds on average to fully denoise a sample.

\section{Additional Baseline Details}
\label{sec:baseline_details}
\paragraph{Prompt Generation.}
Our backbone Motion-I2V~\cite{shi2024motion} supports text-conditioning for image-to-video generation. In the Prompt Generation baseline, we aim to sample diverse and focused object motions using a set of distinct text prompts. To automate this process, we use GPT-4 to generate text prompts that correspond to distinct object motions for a given input image and object. The prompts are then used as text conditioning for Motion-I2V for video generation.

Specifically, we query GPT-4 for the prompts as follows. GPT-4 is first provided the following context: \textit{``I am using a text-based video generator to discover all the different ways a specific object in an image can move, and I wish to generate a set of text prompts in order to achieve this. In particular, I will provide an image and specify an object. For each such specification, I would like to generate 6 text prompts that can be input to the video generator in order to explore the distinct motions the specified object can have in the scene. Remember that we want the motions to be focused only on the specified object and to each be distinct from the other."} We then provide the model with an image along with a text specification of the object in the context of the same conversation to retrieve the text prompts. Some examples of retrieved prompts follow. For a scene with a basketball near a net: \textit{``video of a basketball swishing through the hoop after a jump shot"}, \textit{``video of a basketball bouncing off the rim and falling away from the hoop"}, \textit{``video of a basketball spinning around the rim before dropping in"}. For a scene with a cat on a ledge: \textit{``video of a cat walking gracefully along a ledge with a scenic background"}, \textit{``video of a cat jumping off the ledge gracefully"}, \textit{``video of a cat stopping and looking around curiously"}.

\paragraph{Random Arrows.}
Our backbone Motion-I2V~\cite{shi2024motion} can be conditioned on a drag arrow that describes the rough motion direction and motion magnitude of a point in the image, in an application the authors call \emph{MotionDrag}. In the Random Arrows baseline, we use random drag arrows to explore a diverse set of motions for a selected object.
Specifically, given an object mask $\m$, we set the starting point for the drag arrow to a random point inside the object mask, randomly sample a direction, and sample the length of the drag arrow uniformly from an interval of reasonable lengths ($20$ to $80$ pixels in an image with $320$p resolution). We found that arrow lengths outside this interval tended to either result in zero object motion or implausible motions.

\section{Additional Arrow-based Prompting Details}
\label{sec:snapping_details}
Our arrow-based prompting application shows that Motion Modes can be used to facilitate user interaction with drag-controlled image editors and video generators. As image editors, we work with Drag-A-Part~\cite{li2024dragapart} and DragonDiffusion~\cite{mou2024dragondiffusion}, and as video editors, we use MOFA~\cite{niu2024mofa} and the \emph{MotionDrag} application of Motion-I2V~\cite{shi2024motion}. We take as input a given drag arrow, defined by a start point $\mathbf{a} \in [1, H] \times [1, W]$ and end point $\mathbf{b} \in [1, H] \times [1, W]$, both given as pixel indices for resolution $W \times H$. We then use this drag arrow to retrieve the closest motion $\x$ from our motion set $\X$. Recall that in each frame, our motions describe the same offset of each image point from its starting position as a drag arrow. Thus we can simply compare the drag arrow to each frame of the motion $\x$ at the starting position $\mathbf{a}$ of the drag arrow:
\begin{equation}
    \min_\text{k} \left\lVert\x_{k, \mathbf{a}} - \overrightarrow{\mathbf{ab}}\right\rVert_2,
\end{equation}
where $\x_{k, \mathbf{a}}$ is the offset vector of the motion $\x$ in frame $k$ at the starting point $\mathbf{a}$ of the drag arrow. 
The motion $\x$ with closest distance to the drag arrow describes a motion similar to the drag arrow, but typically has good plausibility and much more detail than the drag arrow. We then convert the retrieved motion back into a representation that the image or video editors can use as input. Specifically, Drag-A-Part can take up to $10$ drag arrows as input, for DragonDiffusion, we can fit up to $100$ arrows into memory, for MOFA, we use up to $50$ arrows (we found that more arrows result in non-static backgrounds), and for Motion-I2V, we can directly provide the retrieved motion $\x$ as conditional input. To convert a motion to $n$ drag arrows, we cluster the offsets in the retrieved frame of the motion into $n$ clusters using K-Means, and use the cluster means as drag arrows.